\documentclass[conference]{IEEEtran}
\IEEEoverridecommandlockouts
\usepackage{cite}
\usepackage{amsmath,amssymb,amsfonts}
\usepackage{algorithmic}
\usepackage{graphicx}
\usepackage{textcomp}
\usepackage{subcaption}
\usepackage{booktabs}
\usepackage{utfsym}
\usepackage{enumitem}
\usepackage{xcolor}
\def\BibTeX{{\rm B\kern-.05em{\sc i\kern-.025em b}\kern-.08em
    T\kern-.1667em\lower.7ex\hbox{E}\kern-.125emX}}

\def\PROG{\textit{PROG}}
\def\Ie{I^{(\epsilon)}}
\def\INCe{\textit{INC}^{(\epsilon)}}
\def\ince{\textit{inc}^{(\epsilon)}}
\def\PROGe{\textit{PROG}^{(\epsilon)}}
\def\spreade{\textit{spread}^{(\epsilon)}}
\def\Gspreade{\textit{GS}^{(\epsilon)}}
\def\cande{{\textit{cand}^{(\epsilon)}}}
\def\S{\textbf{S}}

\newcommand{\constantSet}{\mathcal{C}}

\newcommand{\predicateSet}{\mathcal{P}}
\newcommand{\variableSet}{\mathcal{V}}

\newcommand{\groundLiteralSet}{\mathcal{G}}
\newcommand{\interpretationSet}{\mathcal{I}}
\newcommand{\interpretation}{I}
\newcommand{\semiLattice}{\mathcal{M}}
\newcommand{\delay}{\Delta t}
\newcommand{\satisfactionAtTime}[1]{\models_{#1}}
\newcommand{\program}{\Pi}
\newcommand{\fixpointOperator}{\Gamma}
\newcommand{\groundedLiteral}{g}
\newcommand{\nonGroundedLiteral}{\ell}

\def\avar{\textsf{AVar}}

\begin{document}

\title{Scalable Semantic Non-Markovian Simulation Proxy for Reinforcement Learning}

\author{\IEEEauthorblockN{Kaustuv Mukherji\textsuperscript{\textsection,}\IEEEauthorrefmark{1}, Devendra Parkar\textsuperscript{\textsection}, Lahari Pokala, \\Dyuman Aditya, Paulo Shakarian\IEEEauthorrefmark{2}
\IEEEauthorblockA{\textit{Arizona State University}\\
Tempe, AZ, USA\\
\IEEEauthorrefmark{1}kmukherji@asu.edu, \IEEEauthorrefmark{2}pshak02@asu.edu}
\and
\IEEEauthorblockN{Clark Dorman}
\IEEEauthorblockA{\textit{Scientific Systems Company, Inc.}\\
Woburn, MA, USA\\
clark.dorman@ssci.com}
}
}

\newtheorem{dmodel}{Diffusion Model}[section]
\newtheorem{definition}{Definition}[section]
\newtheorem{example}{Example}[section]
\newtheorem{proposition}{Proposition}[section]
\newtheorem{theorem}{Theorem}[section]
\newtheorem{lemma}[theorem]{Lemma}
\newtheorem{fact}[theorem]{Observation}

\newtheorem{oprob}{Open Problem}[section]
\newtheorem{mex}{Military Example}[section]
\newtheorem{corollary}{Corollary}[section]

\newcommand{\from}[3]{{\bf [{\sc from #1 to #2:} {\small #3}]}}

\def\asetpi{\textsf{aset}_{\Pi(GK)}}
\def\xa{X_{\bV_1}}
\def\xc{X_{\bV_2 - \bV_1}}
\def\xb{X_{\{ v \}}}
\def\cmin{c_{(\min)}}
\def\minsetpi{\textsf{minset}_{\Pi(GK)}}
\def\maxsetpi{\textsf{maxset}_{\Pi(GK)}}
\def\upmaxpi{\textsf{UPMAX}_{\Pi(GK)}}
\def\lomaxpi{\textsf{LOMAX}_{\Pi(GK)}}
\def\minsetpio{\textsf{minset}_{\Pi^{1}(GK)}}
\def\maxsetpio{\textsf{maxset}_{\Pi^{1}(GK)}}
\def\upmaxpio{\textsf{UPMAX}_{\Pi^{1}(GK)}}
\def\lomaxpio{\textsf{LOMAX}_{\Pi^{1}(GK)}}

\def\pio{\Pi^{(1)}}
\def\dppi{DP_{\Pi(GK)}}
\def\dppio{DP_{\pio(GK)}}
\def\vdp{V_{DP}}
\def\edp{E_{DP}}
\def\lbldp{lbl_{DP}}

\def\nset{\textsf{nset}_{GK,pred}}
\def\nsetgoal{\textsf{nset}_{GK,goal}}

\def\asetpio{\textsf{aset}_{\pio(GK)}}

\def\covset{\textsf{covset}_{\pio(GK)}}

\def\md{\textsf{mindist}}
\def\fr{\textsf{fr}}

\def\relu{\mathit{relu}}
\def\incon{\mathit{incon}}
\def\percep{\mathit{percep}}

\def\calr{\mathcal{R}}
\def\calb{\mathcal{B}}
\def\calc{\mathcal{C}}
\def\cala{\mathcal{A}}
\def\calk{\mathcal{K}}
\def\cali{\mathcal{I}}
\def\calt{\mathcal{T}}
\def\calv{\mathcal{V}}

\def\call{\mathcal{L}}

\def\call{\mathcal{L}}
\def\calo{\mathcal{O}}
\def\cald{\mathcal{D}}
\def\cals{\mathcal{S}}
\def\cale{\mathcal{E}}
\def\calf{\mathcal{F}}
\def\calu{\mathcal{U}}
\def\feas{{\mathsf{feas}}}
\def\eqo{\textsf{EQ}_\calo}
\def\eqs{\textsf{EQ}_\cals}
\def\ordo{\textsf{ORD}_\calo}
\def\ords{\textsf{ORD}_\cals}
\def\GK{\textsf{GK}}
\def\FM{{\mathsf{FM}}}
\def\preans{{\mathsf{pre\_ans}}}
\def\answ{{\mathsf{ans}}}
\def\calh{{\mathcal{H}}}

\def\met{\textsf{MET}}

\def\pred{\mathit{pred}}
\def\npred{\mathit{not\_pred}}
\def\cpred{\mathit{cause\_pred}}

\def\rel{\mathit{rel}}
\def\red{\textsf{RED}}
\def\bl{\textsf{BLUE}}
\def\none{\textsf{none}}
\def\rsc{\textsf{RSC}}
\def\pd{\textsf{pd}}
\def\sp{\mathcal{S}}
\def\APT{{\textsf{APT}}}
\def\ISA{{\textsf{ISA}}}
\def\apt{\APT}
\def\calp{\mathcal{P}}
\def\st{\textit{ such that }}
\def\pst{P^{*}}
\def\pstp{P^{*}_{(Pr)}}
\def\pms{\textsf{PM}_\sp}
\def\pc{\textsf{PC}}
\def\mpc{\textsf{mPC}}
\def\setcov{\textsf{SET\_COVER}}
\def\ssa{\textsf{SIM\_SA}}
\def\msa{\textsf{STEADY\_SA}}
\def\NAP{{\textsf{NAP}}}
\def\NAPs{{\textsf{NAPs}}}
\def\avar{\textsf{AVar}}
\newcommand{\tib}[1]{{\textbf{\textit{#1}}}}
\def\T{{\mathbf{T}}}
\def\vp{{\textsf{VP}}}
\def\ep{{\textsf{EP}}}
\def\bG{\textbf{\textsf{G}}}
\def\bV{\textbf{\textsf{V}}}
\def\bE{\textbf{\textsf{E}}}

\def\val{\textit{val}}
\def\inc{\textit{inc}}
\def\incalg{\textit{inc}^{(\textit{alg})}}
\def\incopt{\textit{inc}^{(\textit{opt})}}
\def\incup{\textit{inc}^{(\textit{up})}}
\def\incoptup{\textit{inc}^{(\textit{opt,up})}}
\def\Ialg{I^{(\textit{alg})}}
\def\remaining{\textit{REM}}
\def\INC{\textit{INC}}
\def\PROG{\textit{PROG}}
\def\Ie{I^{(\epsilon)}}
\def\INCe{\textit{INC}^{(\epsilon)}}
\def\ince{\textit{inc}^{(\epsilon)}}
\def\PROGe{\textit{PROG}^{(\epsilon)}}
\def\spreade{\textit{spread}^{(\epsilon)}}
\def\Gspreade{\textit{GS}^{(\epsilon)}}
\def\cande{{\textit{cand}^{(\epsilon)}}}
\def\S{\textbf{S}}

\maketitle
\begingroup\renewcommand\thefootnote{\textsection}
\footnotetext{These authors contributed equally.}
\endgroup

\begin{abstract}
Recent advances in reinforcement learning (RL) have shown much promise across a variety of applications.  However, issues such as scalability, explainability, and Markovian assumptions limit its applicability in certain domains.  We observe that many of these shortcomings emanate from the simulator as opposed to the RL training algorithms themselves.  As such, we propose a semantic proxy for simulation based on a temporal extension to annotated logic. In comparison with two high-fidelity simulators, we show up to three orders of magnitude speed-up while preserving the quality of policy learned. In addition, we show the ability to model and leverage non-Markovian dynamics and instantaneous actions while providing an explainable trace describing the outcomes of the agent actions.
\end{abstract}


\section{Introduction}
\label{sec:introduction}

Recent advances in reinforcement learning (RL) have yielded remarkable progress across various domains, including healthcare~\cite{yu2021reinforcement}, autonomous driving~\cite{kiran2021deep}, and gaming environments such as Atari games~\cite{mnih2015human}. However, scalability concerns hinder RL's capacity to handle complex environments and interactions, while the lack of modularity and portability impedes its adaptability to diverse contexts. Additionally, issues related to explainability, the inherent drawbacks of the Markov assumption, and difficulty implementing safety constraints limit RL's broader applicability in domains demanding rigorous simulation fidelity and reliability. It is crucial to note that the majority of these drawbacks primarily originate from the limitations of the simulation environment employed to train RL agents, rather than intrinsic deficiencies in the underlying RL algorithms themselves. Addressing these challenges necessitates advancements in simulator fidelity and realism. 

In this work, we propose a semantic proxy to replace the simulator based on formal logic.  We show that this approach offers a three order of magnitude speedup over using the native simulation environment.  Further, we train agents in the semantic proxy using standard Deep Q Learning and show that they attain comparable performance to two high-fidelity simulation environments in terms of win-rate and reward.  We also demonstrate advanced capabilities of this framework such as non-Markovian reasoning (which can improve agent performance) as well as how our framework provides an explainable trace of the simulation that is amenable to further symbolic reasoning. The main contributions of this paper are as follows.

\begin{enumerate}[wide, labelindent=0pt]

\item \textit{The introduction of the use of open world temporal logic as a semantic proxy for a simulator.} We show that by using open world temporal logic programming we can successfully create proxies for game environments.  We implemented our approach in PyReason~\cite{pyreason2023} which allows us to leverage a temporal variant of annotated (first order) logic~\cite{shakarian2011annotated, shakarian2022extensions}. The use of a logic program to model a simulation environment is inherently modular and allows direct support for the addition of constraints on agent behavior - without requiring modifications to the RL training regime or reward function. This allows PyReason to leverage abstraction layers (like ROS~\cite{ros2} in robotic applications) to enhance versatility. Similarly, we support adding logic shielding not just within the RL agent like~\cite{alshiekh2018safe, elsayed2021safe}, but also directly within the simulator, detaching it altogether from the RL algorithm and preventing agents from ever executing an unsafe action in any given environment.

\item \textit{We demonstrate a three order of magnitude improvement in runtime over simulation environments while maintaining agent performance.}  The ability to scale while maintaining performance is paramount for accommodating the escalating computational demands of complex environments. PyReason shows up to three orders of magnitude speedup and significantly better memory efficiency over the widely popular simulators Starcraft II (SC2)~\cite{vinyals2017starcraft} and AFSIM~\cite{clive2015advanced}. PyReason-trained policies consistently excel in single-agent and multi-agent scenarios, with less than 10\% reward variance and less than 3\% win rate variance in both SC2 and AFSIM.

\item \textit{We demonstrate that our framework can model non-Markovian and instantaneous actions and that the RL training regime can leverage these capabilities for improved agent performance.} We show that by removing the Markov assumption and by introducing immediate rules in logic, we are able to capture similar environments to real world applications. We illustrate that employing a non-Markovian simulator for training a DQN in a basic wargame context results in a notable 26\% improvement in the win rate compared to adhering to the Markovian assumption.

\item \textit{Our semantic proxy provides a symbolic explainable trace describing the simulation.} Explainability is essential when observing RL simulation outcomes to gain insights into agent decision-making processes and ensure their alignment with intended objectives. PyReason produces fully explainable traces of inference, which can be used in reward shaping and debugging.

\end{enumerate}

The rest of the paper is outlined as follows. In Section~\ref{sec:background} we review our open world temporal logic, which is based on annotated logic~\cite{ks92, shakarian2011annotated} and implemented in PyReason~\cite{pyreason2023} - this is the foundation of the semantic proxy. In Section~\ref{sec:approach} we describe how our semantic proxy replaces the simulator in an otherwise standard reinforcement learning pipeline and point out key extensions to PyReason introduced in this paper to enable this workflow. This is followed by a description of our experimental setup in Section~\ref{sec:setup} to include details on the simulators we examined and the design of each experiment. Section~\ref{sec:results} discusses the results for scalability, portability, non-Markovian dynamics, and explainability. This is followed by a section covering related work (\ref{sec:related}) and thoughts on future work (\ref{sec:future}).

\vspace{3pt}
\noindent\textbf{Codebase:} https://github.com/lab-v2/pyreason-rl-sim

\section{Background}
\label{sec:background}

\vspace{3pt}
\noindent\textbf{Open World Temporal Logic.} We now describe the open world temporal logic we use to build our semantic proxy. For this task, we leverage  Generalized Annotated Logic programs (GAPs) with lower-lattice and temporal extensions from \cite{ks92, shakarian2011annotated, mancalog13, shakarian2022extensions}. The use of GAPs with a lower lattice enables the modeling of open-world scenarios as it allows for the atoms to be associated ``true'', ``false'', or ``no knowledge'' while the temporal extensions are required to model the simulation environments.  Further, the key semantic structure and fixpoint semantics allow for explainable description of the environment's dynamics.

\vspace{3pt}
\noindent\textbf{Syntax.} We consider first order logical language with an infinite set $\constantSet$  of constant symbols, a finite set $\predicateSet$ of predicate symbols, and an infinite set $\variableSet$ of variable symbols. Each predicate symbol $pred \in  \predicateSet$ has an arity. We shall assume that $\calc,\calp,\calv$ are discrete and finite.  In general, we shall use capital letters for variable symbols and lowercase letters for constants.  Similar to previous work~\cite{deepMinIlp2018,thomasNsr21}, we assume that all elements of $\calp$ have an arity of either~1 or~2.

\noindent Atoms and ground atoms are formed in the normal way, e.g. for predicate $pred$, constant $c \in \calc$, and variable $V \in \calv$, $pred(c)$ is a ground atom while $pred(V)$ is a non-ground atom.

\begin{figure}[tb]
    \begin{center}
        \includegraphics[width=\linewidth]{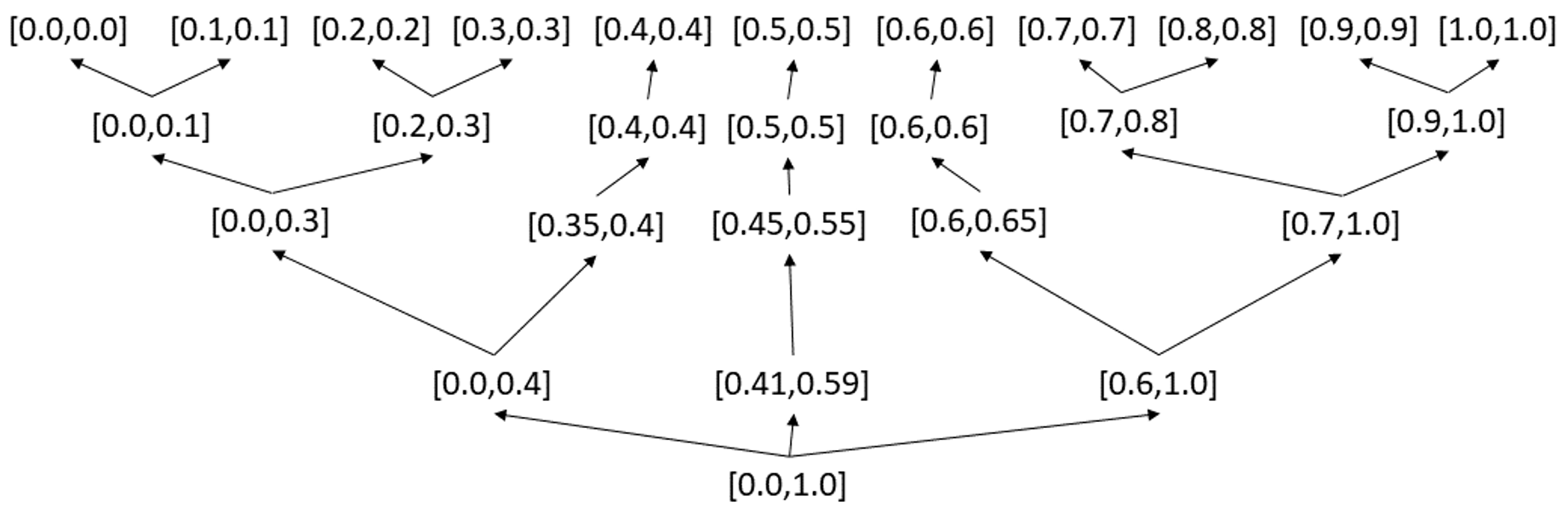}
    \end{center}
    \caption{Example of a lower semi-lattice structure where the elements are intervals in $[0,1]$.}
    \label{fig:lowerLattice}
\end{figure}

\noindent Following~\cite{ks92}, we define a lattice structure $\semiLattice$ where elements consist of subsets of the unit interval where $[0,1]$ (representing total uncertainty) is the lowest element of the lattice while the upper elements are all intervals $[l,u]$ where $l=u$.  The top elements of the lattice include $[1,1]$ (total truth) and $[0,0]$ (total falsehood).  We depict such a lattice in Figure~\ref{fig:lowerLattice}. In annotated logic, atoms are associated with elements of the lattice structure - which is how we enable open-world reasoning (i.e., due to atoms being associated with the bottom lattice).  However, as per~\cite{mancalog13, shakarian2022extensions} we have to extend the definition of the atom to an \textit{annotated atom}.  Given a ground literal $l$ and n element of the lattice $\mu$, $l:\mu$ is an annotated atom.  Functions and variables are also permitted in the annotations (see \cite{mancalog13, shakarian2022extensions} for further details).

\vspace{3pt}
\noindent We propose a modified version on GAP rule definied in \cite{shakarian2022extensions}:  
\noindent\begin{definition}[GAP Rule]
If $\nonGroundedLiteral_0:\mu_0, \nonGroundedLiteral_1:\mu_1,\ldots,\nonGroundedLiteral_m:\mu_m$ are annotated literals (such that for all $i,j \in 1,...,m$, $\nonGroundedLiteral_i\not\equiv \nonGroundedLiteral_j$), then
\begin{align*}
r\equiv \nonGroundedLiteral_0:\mu_0 &\xleftarrow[\delay]{} \nonGroundedLiteral_1:\mu_1\,\wedge\,\ldots\wedge\, \nonGroundedLiteral_m:\mu_m \tag*{\llap{\text{$\delay \geq 0$}}}
\end{align*}
is called a \emph{GAP rule}.  We will use the notations $head(r)$, $delay(r)$ and $body(r)$ to denote $\nonGroundedLiteral_0$, $\delay$ and $\{\nonGroundedLiteral_1,\ldots,\nonGroundedLiteral_m\}$ respectively. When $m=0$ ($body(r)=\emptyset$), the above GAP-rule is called a \emph{fact}.  A GAP-rule is \emph{ground} iff there are no occurrences of variables from $\variableSet$ in it. $\delay$ is the temporal gap between when the rule is fired and when it's effects are applied. If body(r) is satisfied at time t, then the annotation of $\nonGroundedLiteral_0$ changes to $\mu_0$ at time $t+\delay$. A temporal logic program $\Pi$ is a finite set of GAP rules.
\end{definition}

Our key intuition is that a program $\Pi$ can be used to capture the dynamics of an environment.  In practice, a program $\Pi$ is comparable to code written in languages like PROLOG allowing for flexible environmental definitions that can align precisely with constructs in a simulation.  We provide an example in Table~\ref{tab:example_rules}.  We can think of a program as consisting of two subsets of rules: one dictating the dynamics of the environment and the other dictating agent actions.  The former would be generated as part of the game design while the later can be the policy produced by a reinforcement learning algorithm.

\vspace{3pt}
\noindent\textbf{Semantic Interpretation.}  An annotated logic program $\Pi$ is associated with a semantic interpretation that maps literal-time point pairs to annotations.  Our intuition is that this structure, which is produced as output of deductive inference can directly describe the change in the environment resulting from a set of rules and an agent's actions.  Notably, the interpretation is entirely symbolic, hence fully explainable in terms of the logical language.  We provide a formal definition of an interpretation and associated satisfaction relationship below.

\vspace{3pt}
\noindent\begin{definition}[Semantic Interpretation]\label{def:interp}
Let us assume a sequence of timepoints $T = t_1, ..., t_{max}$. Then, an interpretation $\interpretation$ is any mapping  $ \groundLiteralSet \times T \to \semiLattice$ such that for all literals~$l$, we have $\interpretation(l, t) = \neg(\interpretation(\neg l, t))$.  Here, $\groundLiteralSet$ is the set of all ground literals. The set~$\interpretationSet$ of all interpretations can be partially ordered via the ordering: $\interpretation_1\preceq \interpretation_2$ iff for all ground literals $\groundedLiteral \in \groundLiteralSet$ and time t, $\interpretation_1(\groundedLiteral, t)\sqsubseteq \interpretation_2(\groundedLiteral, t)$. $\interpretationSet$ forms a complete lattice under the $\preceq$ ordering.
\end{definition}

\vspace{3pt}
\noindent\begin{definition}[Satisfaction for annotated ground literal]
An interpretation $\interpretation$ at time $t$ \emph{satisfies} annotated ground literal $\groundedLiteral:\mu$, denoted $ \interpretation \satisfactionAtTime{t} \groundedLiteral:\mu$, iff $\mu \sqsubseteq \interpretation( \groundedLiteral, t)$. 
\end{definition}

\vspace{3pt}
\noindent\begin{definition}[Satisfaction of GAP rule]
$\interpretation$ satisfies the ground GAP-rule
\begin{eqnarray*}
r\equiv \groundedLiteral_0: \mu_0 &\xleftarrow[\delay]{} & \groundedLiteral_1:\mu_1\wedge\,\ldots\,\wedge\, \groundedLiteral_m:\mu_m
\end{eqnarray*}
denoted $\interpretation \models r$, iff for $t \leq t_{max} - \delay$ where for all $\groundedLiteral_i: \mu_i \in $ body(r), if $\interpretation \satisfactionAtTime{t} \groundedLiteral_i:\mu_i$ then $\interpretation \satisfactionAtTime{t + \delay} $ head(r). $\interpretation$ satisfies a non-ground literal or rule iff $\interpretation$ satisfies all ground instances of it.
\end{definition}

\vspace{3pt}
\noindent\textbf{Fixpoint-based Inference in Annoated Logic.}
In~\cite{mancalog13, shakarian2022extensions} the authors present a fixpoint operator for identifying the logical outcome of a logic program.  Our intuition is that the fixpoint operator essentially performs a simulation - all the while recording the changes.  We note that under the assumption of consistency, this operator produces an exact result in polynomial time (see Theorem 3.2 and 3.4 of \cite{shakarian2011annotated}) and recent implementation provides practical speed-ups and consistency checking while maintaining these guarantees~\cite{pyreason2023}.  We define it formally below:

\vspace{3pt}
\noindent\begin{definition}[Fixpoint operator]
Suppose $\program$ is any GAP and $\interpretation$ an interpretation. The fixpoint operator $\fixpointOperator$ is a map from interpretations to interpretations and is defined as
\[
\fixpointOperator(I)(\groundedLiteral_0, t) = \mathbf{sup}(annoSet_{\program,\interpretation}(\groundedLiteral_0, t)),
\]
where $annoSet_{\program,\interpretation}(\groundedLiteral_0, t) = \{\interpretation(\groundedLiteral_0, t)\}\cup
\{\mu_0$ such that for all ground rules r $\in \program$, where head(r)=$\groundedLiteral_0:\mu_0$, for all $\groundedLiteral_i: \mu_i \in$ body(r) and delay(r) $\leq t$ and $\interpretation \satisfactionAtTime{{t-delay(r)}} \groundedLiteral_i:\mu_i$\}. 
Here delay(r) is the delay associated with specific rule r.

\vspace{3pt}
\noindent Given natural number $i > 0$, interpretation $\interpretation$, and program $\program$, we define $\fixpointOperator^i(I)$, then multiple applications
of $\fixpointOperator$:\\
$\fixpointOperator^i(I) = \fixpointOperator(I)$ if $i=1$ and $\fixpointOperator^i(I) = \fixpointOperator(\fixpointOperator^{i-1}(I))$ otherwise.
\end{definition}

\noindent We note that the fixpoint operator maps \textit{all} time-point-literal pairs to time-point literal pairs - so essentially revising the entire sequence of timepoints at once.  This contrasts with approaches such as MDPs which produce a new state at each time-point.  This allows for direct modeling of non-Markovian dynamics in the framework.

\section{Approach}
\label{sec:approach}
In this section we detail our approach to using logic as a simulator and describe PyReason, our software implementation. Then we introduce new enhancements to PyReason, including the ability to interface with RL agents.

\vspace{3pt}
\noindent\textbf{Logic as simulator for Reinforcement Learning.} Deep Reinforcement Learning (RL) algorithms typically require a simulator to learn an agent policy. However, traditional simulators have several drawbacks like speed and data efficiency, lack of explainability and modularity, inextensibility without retraining. We propose annotated logic (implemented in PyReason) to address these issues and compare it with some well established simulation environments.

\vspace{3pt}
\noindent\textbf{The PyReason software\footnote{PyReason github: https://github.com/lab-v2/pyreason}  (Recap of prior work).}
PyReason~\cite{pyreason2023} offers a comprehensive and flexible framework for reasoning based on generalized annotated logic. It supports various extensions, including temporal, graphical, and uncertainty-related features, which enable the capture of a wide range of logics, such as fuzzy, real-valued, interval, and temporal logics.

Built on modern Python, PyReason is specifically designed to handle graph-based data structures efficiently, making it compatible with data exported from popular graph databases like Neo4j and GraphML.

The core of PyReason lies in its rule-based reasoning, which enables handling uncertainty, open-world novelty, non-ground rules, quantification, and other diverse requirements seamlessly. The system remains agnostic to the selection of t-norm, providing flexibility in utilizing different logical connectives.

One of the key strengths of PyReason is its speed and machine-level optimized fixpoint-based deduction approach. This ensures efficient and scalable reasoning capabilities, even when dealing with large graphs with over 30 million edges. Consequently, PyReason facilitates explainable AI reasoning, providing valuable insights into the decision-making process and the logic behind reaching specific conclusions.

Our description of the world as a knowledge graph (KG) is notable as it adds support to applications where a policy must be learnt via reasoning over context related KGs such as~\cite{XR4DRAMA23}. Additionally, recent progress in developing Knowledge Graphs (KGs) for probabilistic reasoning, as demonstrated by studies such as~\cite{vidKG23, ontologyKG23, DocSemMap22}, highlight the potential role of our framework in a wide range of practical applications.

The logic based approached used in PyReason is also inherently modular allowing for independently trained or created components. Finally, the logic in PyReason can be extended simply by adding symbols to an existing logic program.

\vspace{3pt}
\noindent\textbf{Immediate Rules (New in this paper).}
We introduce a feature called immediate rules. Immediate rules are applied immediately and make the program search for new applicable rules whose clauses might now be satisfied because of the immediate rule. Previously it was impossible for two rules with the same ${\Delta}t$ to influence each other. This is required when the shooting action (see Section~\ref{sec:setup}) is brought into the picture because there are multiple events occurring with the same ${\Delta}t$ but they're all interconnected. We note that this is possible without any extensions to annotated logic as the temporal extensions we use (based on~\cite{mancalog13, shakarian2011annotated, pyreason2023}) have no requirement that two time units be uniformly separated in actual time.

\vspace{3pt}
\noindent\textbf{Implementation Improvements.}
For this work, we also modified various aspects of PyReason to improve memory management particularly to better support the analysis of graphical structures representing geospatial areas as well as generally mature the software.

\vspace{3pt}
\noindent\textbf{Interfacing with an RL agent (New in this paper).}
We introduce PyReason-gym\footnote{PyReason-gym github: https://github.com/lab-v2/pyreason-gym}, an OpenAI Gym wrapper that allows easy interfacing with a grid world that uses PyReason as the simulation and dynamics engine. We use logical rules to dictate how the agents move around through the grid world and how bullets and obstacles interact with the agents. RL agents can use our gym environment as a simulator, as action(s) chosen by agent's policy is processed and world state and reward(s) are returned by PyReason-gym. It also has the capability of outputting a trace of all events that happened and when they happened because that is a core functionality of PyReason. PyReason-gym has several settings that allow it to be very efficient and consume a constant amount of memory.

\section{Experimental Setup}
\label{sec:setup}
In this section we introduce the two popular simulators we benchmark our approach against, outline the two game scenarios we use in our experiments, analyze the limitations of Markov assumptions and discuss the RL training methodology adopted.

\vspace{3pt}
\noindent\textbf{Popular Simulators.}
To justify PyReason to be an appropriate simulator, we must first compare it to established simulators in the field. For this we choose two simulators:
\begin{enumerate}[wide, labelindent=0pt]
    \item \textbf{Starcraft II} (SC2) is a popular real-time strategy (RTS) video game developed by Blizzard Entertainment and has a competitive multiplayer aspect that involves managing resources, building armies, and engaging in tactical battles. Due to its complex gameplay and emphasis on strategic decision-making, it has been considered as a potential tool for military simulations. We extended Deepmind's PySC2~\cite{vinyals2017starcraft} to utilize the Starcraft II environment in our experiments\footnote{Extensions to PySC2: https://github.com/lab-v2/pysc2-labv2}.
    \item \textbf{Advanced Framework for Simulation, Integration, and Modeling software} (AFSIM)~\cite{clive2015advanced} is a powerful simulation tool used by the United States Department of Defense (DoD) for various purposes, including training, analysis, experimentation, and mission planning. AFSIM is developed by the Air Force Research Laboratory (AFRL) and is utilized primarily by the United States Air Force (USAF) as well as other branches of the military and defense organizations. AFSIM is a high-fidelity modeling and simulation software designed to provide realistic representations of aerial warfare scenarios and environments. It enables the USAF to assess and analyze the performance of various systems, strategies, and tactics in simulated combat situations.
\end{enumerate}

\noindent In order to compare PyReason with SC2 and AFSIM, we design the scenarios and game dynamics in all three simulators.

\begin{figure}[tb]
    \begin{center}
        \includegraphics[width=0.75\columnwidth]{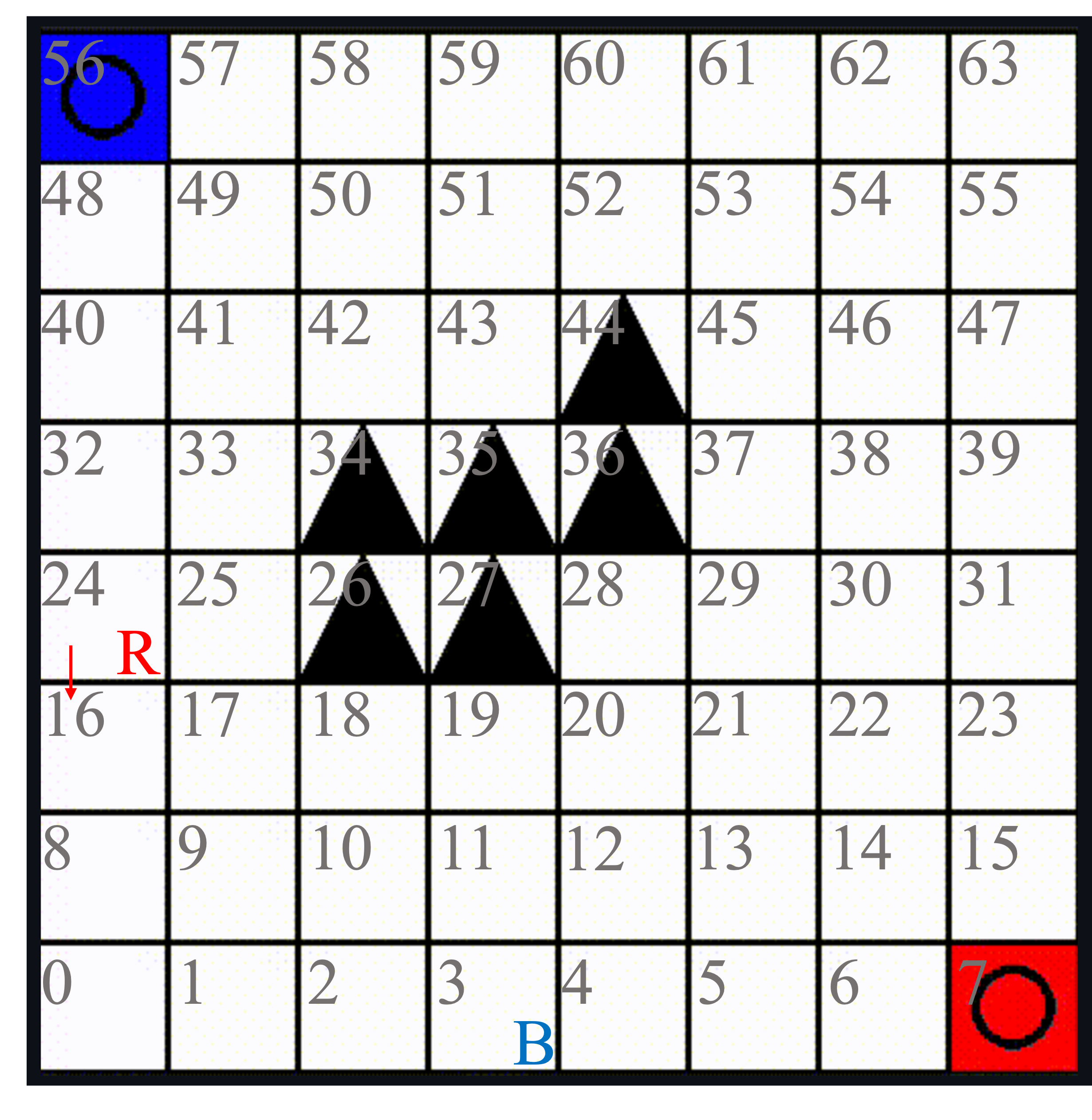}
    \end{center}
    \caption{Grid map for the scenario. Red (bottom-right) and Blue (top-left) squares are fixed base locations for each team. All agents start at their respective base locations. Obstacles (mountains) are shown with black triangles. Bottom left quadrant of the grid map is marked with indices, to aid the understanding of the explainable trace in Table~\ref{tab:short_rule_trace}.}
    \label{fig:game-setup}
\end{figure}

\vspace{3pt}
\noindent\textbf{Game Setup.} We design a simple grid world war game as shown in Fig.~\ref{fig:game-setup}. The basic scenario has two teams (red and blue) of one agent each. Each team has a base, and there are also a few obstacles (shown as mountains) in the environment which are impenetrable and impassable. For this base scenario, the objective of the game is to capture (reach) the rival base before the enemy can do the same. The red team follows our learnt RL policy (the agent(s)), whereas the blue team follows a pre-defined base policy (the opponent(s)) described later in this section. Later on we build upon this basic scenario by adding more agents and then extending the action and observation spaces.

\vspace{3pt}
\noindent\textbf{Comparison with baseline simulation environments.}
Allowing the agents to take random actions in the grid world, we compare the scaling capability of our software against other simulators by comparing the runtime and memory utilization over a large number of actions for different number of agents per team.

\noindent Next we wanted to verify if a Reinforcement Learning (RL) agent trained in PyReason (PR) can provide comparable performance to AFSIM (AFS) and PySC2 (SC2). For this, we considered two cases: single agent and multi (five) agents per team. At certain intervals during the training process, policies were extracted and were used to play the base scenario described earlier 500 times in each of the three simulators (PyReason, AFSIM, and PySC2) and the outcomes were compared.

\vspace{3pt}
\noindent\textbf{Extending the action space with shooting in PyReason.} Some simulations (e.g., Starcraft II) do not separate movement and shooting (i.e., the agent always shoots when in line of sight with an enemy). This however, is undesirable in any military sim looking to emulate real battlefield scenarios. Strategies are often pragmatic, with shooting often limited and highly tactical. Practical issues such as limited ammunition and avoiding exposure are important considerations here. Hence, we build upon the basic scenario by integrating shooting into PyReason, independent from movement actions - allowing RL agents to learn varied and in-depth strategies - and in the process ensuring our implementation fits our eventual goal of a faithful miliary simulation. For this advanced scenario, each agent is provided with three bullets and at each timepoint they may either choose to move or shoot. They may also choose to not take any action. Other than capturing the enemy base, a team can win by eliminating all enemy agents.

\vspace{3pt}
\noindent\textbf{Learning policies with RL.} Our approach is agnostic to any specific RL algorithm. Hence for this work we chose to use the widely popular and versatile Deep Q learning (DQN) algorithm~\cite{mnih2015human} for all of our experiments. Based on a specific application or domain, a suitable algorithm can be seamlessly used in place of DQN. In our implementation, we combine a shallow Q-Net architecture with techniques discussed in~\cite{mnih2015human} such as experience replay, stable learning and hard updates for target network. In our architecture we use one hidden layer between input and output layers; 64 state variables (one for each grid cell) and an action space of 5 (for base scenario) or 9 (for advanced scenario).
\noindent Observation state space available to the agent was symbolic in nature and its size varied between experimental setups as follows:
\begin{enumerate}[wide, labelindent=0pt, label=(\roman*)]
    \item Four for single agent in the base scenario. Two each for the current positions of the agent and the opponent.
    \item Seven for single agent in the advanced scenario. One for the number of opponent bullets in the environment, two for the nearest bullet position, in addition to, two each for the current positions of the agent and the opponent.
\end{enumerate}
For multi-agent setups, the observation space is multiplied by the number of agents in each team. For the special non-Markovian setup described later, observation space is doubled as observations from previous timestep are considered. For experiments in multi-agent environments we learn non co-operative single agent policies using multi-agent sampling. We use the widely adopted Smooth L1 loss function, instead gradient clipping as described in the seminal DQN work.

\noindent We use the following reward function (rewards related to shooting actions only applicable for the advanced scenario):
\label{sec:reward_fn}
\begin{enumerate}[wide, labelindent=0pt, label=(\roman*)]
    \item Terminal state rewards: +250 for a win, -250 for a loss, +400 for shooting an opponent, -200 for getting shot.
    \item Non-terminal state rewards: -2 for a valid action, -200 for an unsafe or illegal action, -10 for an invalid action such as trying to shoot after exhausting ammunition.
\end{enumerate}

\noindent We define the behavior of the opponent using a stochastic base policy. At each timestep it tries to move closer to the enemy base by reducing the manhattan distance with a probability of 0.7, or chooses a random action from the action space with a probability of 0.3. In the advanced scenario, shooting is prioritized over movement until ammo is exhausted.

All RL policies described in this paper were learnt on a NVIDIA A100 GPU with 80GB memory and 40 cores of AMD EPYC 7413 with 378GB memory.

\vspace{3pt}
\noindent\textbf{Shielding in RL.} As discussed in Section~\ref{sec:introduction}, we incorporate logic shielding within the reward function, as well as, the simulation environment itself. In the reward function, the agent is heavily penalized for taking an unsafe action, such as, trying to move through the mountains or choosing an action that takes it out of bounds of the map. While this approach encourages the agent to learn policies that avoid unsafe actions, it provides no guarantees. Adding shielding in the simulator itself ensures that even if the agent was to choose an unsafe action, our rule based environment dynamics can detect and stop the execution of such actions in runtime. Furthermore, we can leverage these dynamics to prevent illegal actions such as, shooting when ammo has already been exhausted.

\vspace{3pt}
\noindent\textbf{Exploring limitations of Markov assumption.}
The Markov assumption in RL is the assumption that the future state of an agent only depends on its current state and action, and not on the history of states and actions that led to the current state. As this simplifies the problem and enables the use of techniques like Markov Decision Processes (MDPs) and the Bellman equation, many well established simulators make this assumption. However, many real-world environments are not truly Markovian. In some cases, the current state may not contain all the relevant information for decision-making. This is especially important for simulators replicating realistic military combat environments where various key factors like logistical support, conflict history, long-term intelligence data, patterns in surveillance reports - that go into tactical decision making are non-Markovian in nature.

PyReason does not make a Markov assumption and we exhibit it's capability by creating a simple experiment with non-Markovian dynamics. We consider a two-agents per team, advanced scenario as described earlier. We introduce a modification to one agent within each team, constraining its ability to execute actions to once every two timesteps, with the added stipulation that each of its movement actions require two timesteps to complete. We learn to play the game in two different ways. In the initial approach, the player adheres to a Markov assumption, utilizing solely the current state information. Conversely, in the second approach, the player gains access not only to the present state data but also to observations from the preceding time step. We compare the success of the two methods by evaluating learnt policies over 500 games after every 32,000 training epochs.

\section{Results}
\label{sec:results}
In this section, we present experimental results comparing our approach's scalability in Starcraft II and AFSIM, along with its ability to learn policies in PyReason and port them to other simulators. We then explore whether incorporating non-Markovian dynamics in the simulation can improve RL algorithms' ability to learn effective policies for complex games. Additionally, we demonstrate the explainability of our approach using a rule trace, highlighting its potential in reward shaping during training.

\begin{figure}[tb]
    \begin{center}
        \begin{subfigure}{0.49\columnwidth}
            \begin{center}
                \includegraphics[width=\linewidth, trim=2 2 2 2, clip]{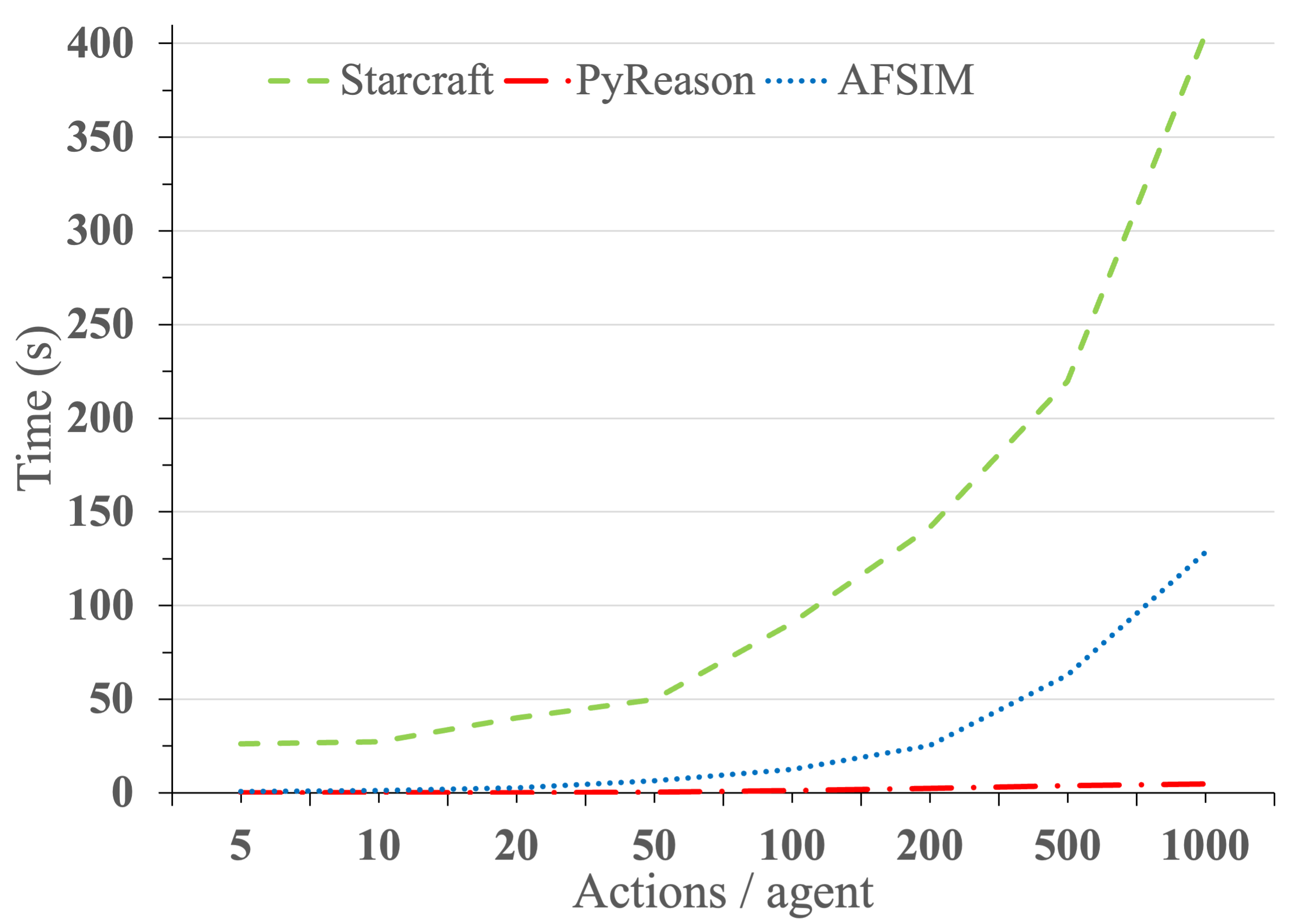}
            \end{center}
        \end{subfigure}
        \begin{subfigure}{0.49\columnwidth}
            \begin{center}
                \includegraphics[width=\linewidth, trim=2 2 2 2, clip]{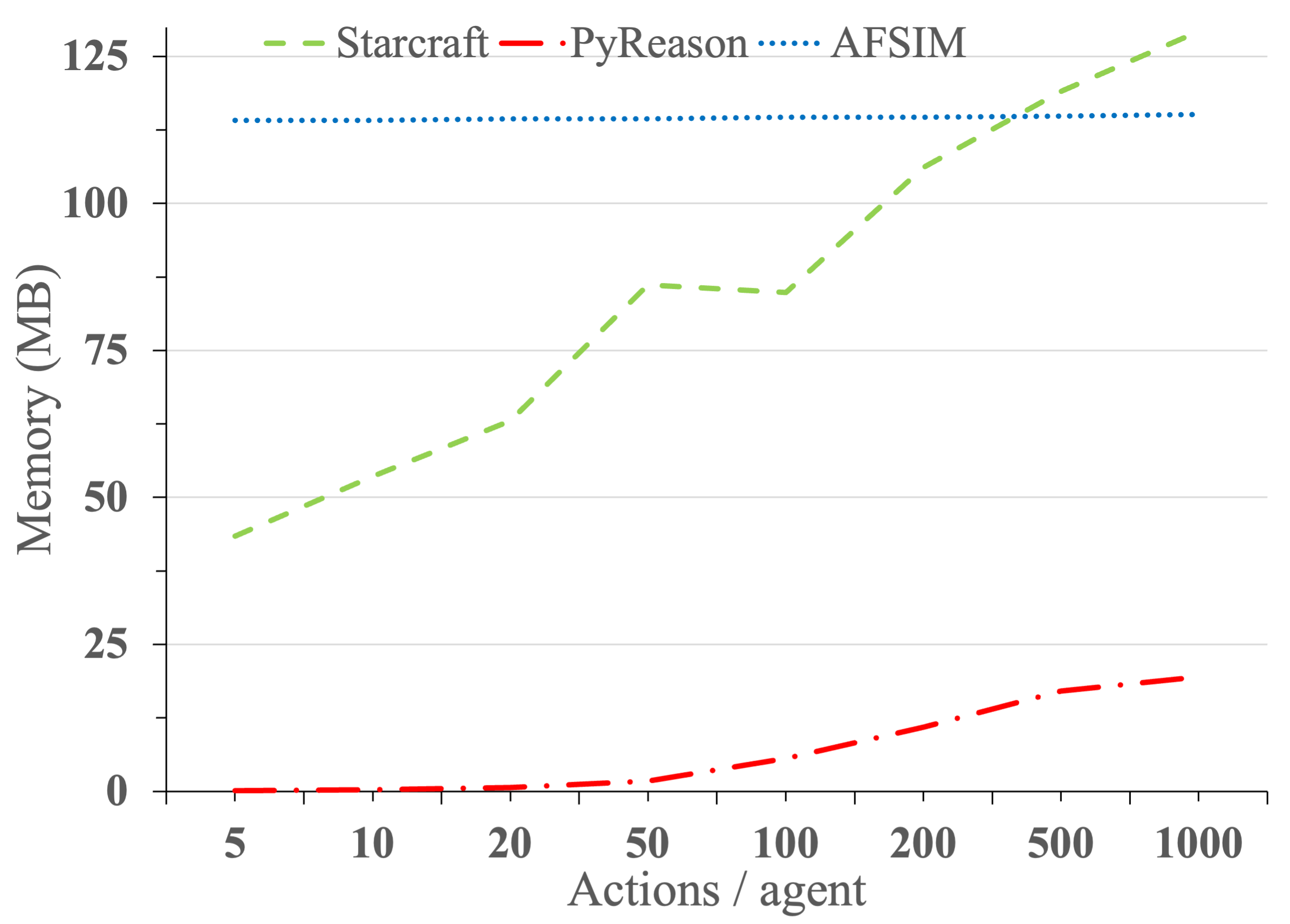}
            \end{center}
        \end{subfigure}
        \begin{subfigure}{0.49\columnwidth}
            \begin{center}
                \includegraphics[width=\linewidth, trim=2 2 2 2, clip]{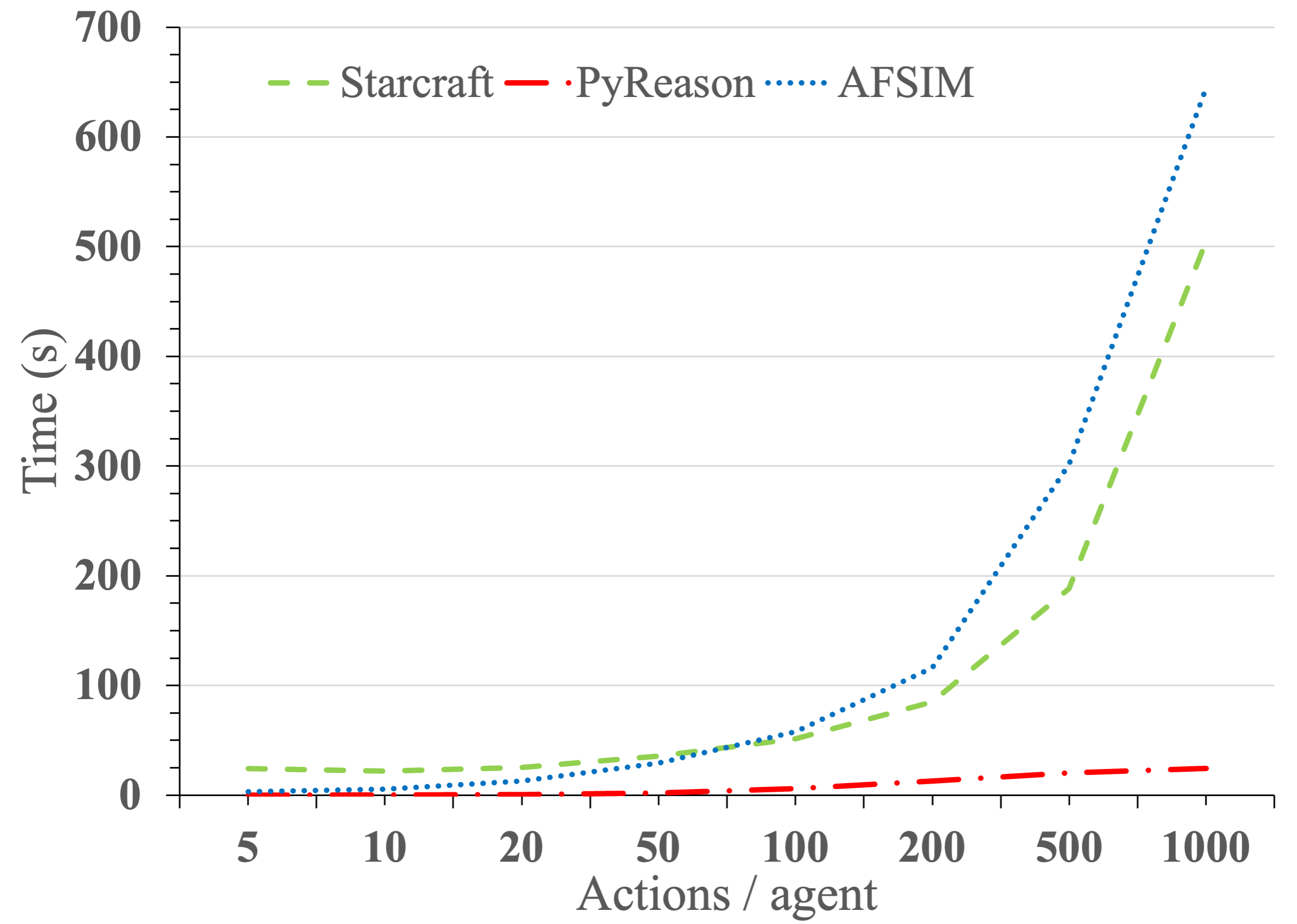}
            \end{center}
        \end{subfigure}
        \begin{subfigure}{0.49\columnwidth}
            \begin{center}
                \includegraphics[width=\linewidth, trim=2 2 2 2, clip]{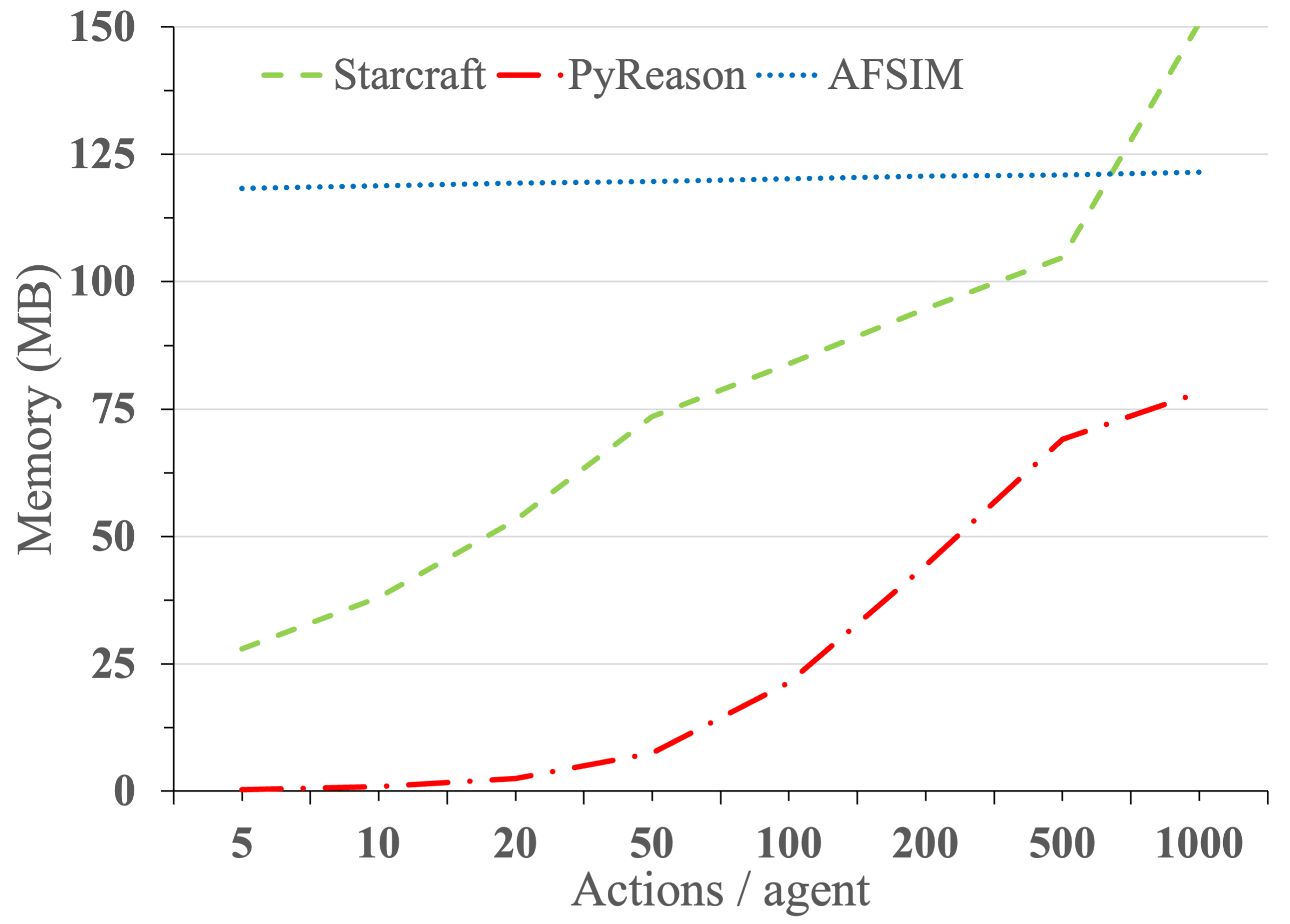}
            \end{center}
        \end{subfigure}
    \end{center}
    \caption{Runtime and Memory utilization comparison when 5 (top), 20 (bottom) agents/team take random actions in three sim environments.}
    \label{fig:runtime-mem-comp}
\end{figure}

\vspace{3pt}
\noindent\textbf{Scalability.}
Figure~\ref{fig:runtime-mem-comp} show the scaling capability of different simulators tested. The experiments were performed on an AWS EC2 container with 96 vCPUs (48 cores) and 384GB memory. We noted that, among the two established simulation environments, AFSIM generally performed better with 5 agents per team.  With 20 agents per team, AFSIM is overtaken by SC2 as the actions per agent increases.  This would be expected as AFSIM is designed as a high-fidelity simulation environment, so we would expect greater computational cost with more complex situations. PyReason consistently outperformed SC2, achieving anywhere from a one to nearly three orders of magnitude improvement.  Though PyReason performs comparably to AFSIM for lower actions per agent (which are arguably the least important in practice), it also achieved comparable multiple order-of-magnitude improvement in terms of runtime as the number of actions per agent increased. This suggests that PyReason will scale to large environments where the traditional use of simulators would otherwise prohibit model training.

Additionally, we examined memory consumption (Figure~\ref{fig:runtime-mem-comp}). PyReason uses considerably lower memory over SC2 over all configurations while having sub-linear ($R^2 = .84$) growth with action and agent space. AFSIM's strength as a large-scale military simulator is shown here with little effect on memory consumption with change in agents or actions, however it has a large base memory cost which was still significantly higher than that of PyReason for the largest case considered (40,000 actions in total).

\vspace{3pt}
\noindent\textbf{Portability.}
When policies learnt in PyReason played the base scenario, comparable numbers were observed for all three simulators as shown in Table~\ref{tab:portability}. Variance can be attributed to inherent randomness in learnt policies. These results suggest that the approach is generalizable as an agent trained in PyReason can be ported to various simulation environments and achieve comparable reward and win percentage.

\begin{table}[tb]
\caption{Performance metrics when PyReason trained policies were used to play the game on different simulators for single and multi (5) agent scenarios (numbers in parentheses specify difference from PyReason).}
\label{tab:portability}
\begin{tabular}{@{}p{0.00001\textwidth}p{0.027\textwidth}cccccc}
\toprule
 \# & Epochs & \multicolumn{3}{c}{Avg. Reward} & \multicolumn{3}{c}{Win \%} \\ \cmidrule(l){3-5} \cmidrule(l){6-8} 
 & & PR & SC2 & AFS & PR & SC2 & AFS \\ \midrule
1 & 400K & -209.87 & -210.15 & -222.65 & 0.0 & 0.0 & 0.0 \\
 &  &  & (-0.13\%) & (-6.09\%) & & (0) & (0)\\
 & 544K & 162.51 & 165.64 & 168.04 & 43.0 & 42.8 & 44.0 \\
 &  &  & (+1.93\%) & (+3.40\%) &  & (-0.2) & (+1)\\
 & 760K & 482.50 & 487.00 & 473.50 & 97.6 & 100.0 & 100.0 \\
 &  &  & (+0.93\%) & (-1.87\%) &  & (+2.4) & (+2.4) \\ \midrule
5 & 112K & -913.27 & -986.88 & -880.16 & 0.0 & 0.0 & 0.0 \\
&  &  & (-8.06\%) & (+3.63\%) &  & (0) & (0) \\
& 352K & -5166.99 & -5548.18 & -5229.43 & 1.6 & 1.8 & 0.0 \\
 &  &  & (-7.38\%) & (-1.21\%) &  & (+0.2) & (-1.6) \\
 & 1536K & 1899.71 & 1860.05 & 1765.43 & 79.4 & 78.8 & 79.0 \\
 &  &  & (-2.09\%) & (-7.07\%) &  & (-0.6) & (-0.4) \\ \bottomrule
\end{tabular}
\end{table}

\begin{figure}[tb]
    \begin{center}
        \includegraphics[width=0.8\columnwidth, trim=2 2 2 2, clip]{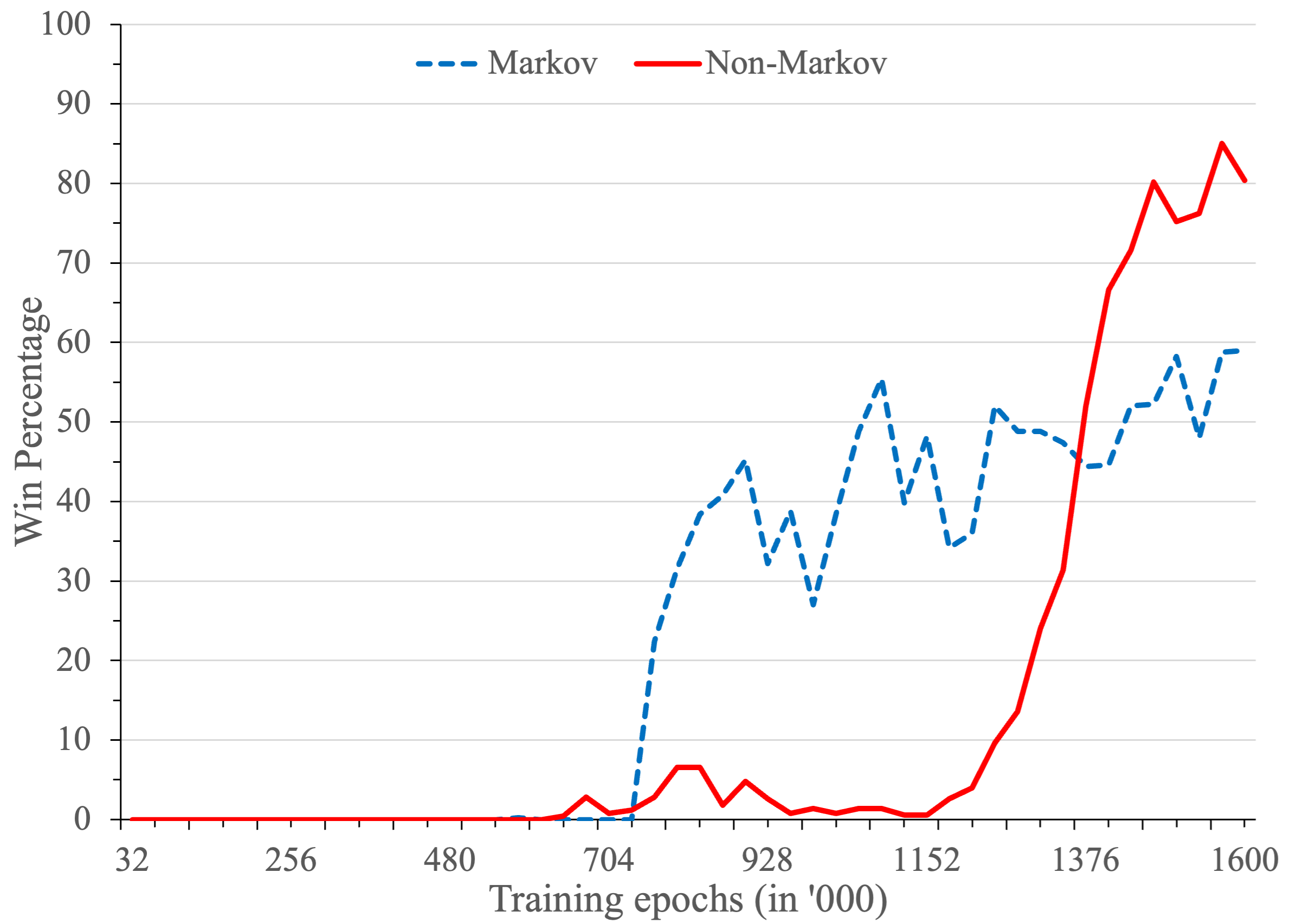}
    \end{center}
    \caption{Win percentage for policies learnt with Markovian and non-Markovian dynamics.}
    \label{fig:shoot-markov-non-markov-comp}
\end{figure}

\vspace{3pt}
\noindent\textbf{Non-Markovian Dynamics.} Evolution of the performance of policies learnt with and without the Markov assumption is shown in Fig.~\ref{fig:shoot-markov-non-markov-comp}. Both agents underwent training for a duration of up to 1.6 million epochs, with policy evaluations conducted at intervals of 32,000 epochs. Each policy was used to play the advanced scenario 500 times to obtain a win percentage. Evaluations were carried out on 48 cores of AMD EPYC 7413 with 378GB memory. Markovian policies obtained a peak performance of 59\%, significantly lower than 85\% achieved by the non-Markovian policies. However we observe that policies learnt in Markovian setting attained decent performance with noticeably less training, which is unsurprising given the doubling of the observation space in the non-Markovian case. When examining the most effective policy within each category, the removal of the Markov assumption resulted in an increase in the average number of actions per agent required to secure a single victory, rising from 15.51 to 18.01. This observation suggests the acquisition of a policy characterized by greater complexity, yet one that exhibits enhanced reliability.
Despite the relative simplicity of our experiment, a noteworthy performance enhancement was observed. This underscores the essentiality and significance of accommodating non-Markovian dynamics within simulation environments.

Win percentage over 500 trials, for policies learnt with non-Markovian dynamics is shown in Fig.~\ref{fig:shoot-markov-non-markov-comp}. Each team is made up of 1 fast-moving and 1 slow-moving agent. Action space is extended to include two timesteps.

\begin{table*}[tb]
\caption{Example rules in first order logic and descriptions in natural language.}
\label{tab:example_rules}
\begin{tabular}{@{}p{0.1\textwidth}p{0.42\textwidth}p{0.43\textwidth}}
\toprule
Rule Identifier & Rule & Natural Language \\
\midrule
m\_Down\_on & \(moveDown(A):[1,1] \leftarrow_{\Delta t=0} agent(A):[1,1] \wedge moveDir(A, down):[1,1] \wedge atLoc(A, X):[1,1] \wedge downLoc(Y, X):[1,1] \wedge blocked(Y):[0, 0]\) & If $A$ is an agent (annotated $[1,1]$) at location $X$, chooses to move in downward direction to $Y$ which is not blocked, then interpretation(label) $moveDown(A)$ is updated to $[1,1]$.\\
\vspace{0.5pt} &\vspace{0.5pt} &\vspace{0.5pt} \\
s\_Left\_on & \(shootLeftB(A):[1,1] \leftarrow_{\Delta t=0} agent(A):[1,1] \wedge team(A, blue):[1,1] \wedge health(A):[0.1,1] \wedge ammo(A):[0.1,1]  \wedge shootLeft(A):[1,1] \) & If $A$ is an agent of the blue team, chooses to shoot left, then label $shootLeftB(A)$ is updated to $[1,1]$ iff $A$ has non-zero health and remaining ammo.\\
\bottomrule
\end{tabular}
\end{table*}

\vspace{3pt}
\noindent\textbf{Explainability.} A major drawback of Deep learning based systems is the lack of any semantic understanding of the output. Logic programs inherently support semantic understanding. PyReason reasons over graphs using first order logical rules (an example is shown in Table~\ref{tab:example_rules}) and produces a explainable trace detailing rules fired at different timesteps, constants used for grounding and interpretation changes. The explainable trace is a direct result of the semantic structure of logic. This makes our approach completely explainable and allows the user to understand system behavior and helps in debugging errors. 

Two examples of how we leveraged this to improve our reward function given in section~\ref{sec:reward_fn} are:
\begin{enumerate}[wide, labelindent=0pt, label=(\roman*)]
    \item Initially we had set the penalty for getting shot at 400. However, from rule traces we observed that the agent was learning to prioritize hiding behind impenetrable mountains and take a safety first approach, instead of trying to win the game. Halving the penalty to 200 produced a more balanced policy.
    \item The penalty for trying to shoot after exhausting ammunition was set to a minor value of 10 after observing that higher values led to the agent avoiding shooting altogether.
\end{enumerate}

An excerpt of a rule trace is shown in Table~\ref{tab:short_rule_trace}. Excerpt shown begins at timestep 16 of one of our experiments. Initial conditions are as depicted in Fig.~\ref{fig:game-setup}. `R' and `B' respectively show the location of the red and blue agents at the beginning of this example. As the red agent moves downward from it's starting location (from `24' to `0' through `16' and `8'), the blue agent decides to shoot to the left so as to intercept red (at `0'). However, red has seemingly learnt to predict the bullet path and evade it. So it backtracks (to `16').

Rule \textbf{m\_Down\_on} presented in Table~\ref{tab:example_rules} is fired at timestep 16 (and also, 17, 18) and is pictorially shown with a red arrow in Fig.~\ref{fig:game-setup} and in bold in Table~\ref{tab:short_rule_trace}.

\begin{table}[tb]
\caption{An extract of a rule trace produced by the PyReason software.}
\label{tab:short_rule_trace}
\begin{tabular}{p{0.001\textwidth}p{0.117\textwidth}p{0.055\textwidth}p{0.04\textwidth}p{0.04\textwidth}p{0.098\textwidth}}
\toprule
t & Node/Edge & Label & Old Bound & New Bound & Rule fired \\ \midrule
0 & 26 & blocked & {[}0.0,1.0{]} & {[}1.0,1.0{]} & - \\
0 & 27 & blocked & {[}0.0,1.0{]} & {[}1.0,1.0{]} & - \\ \midrule
\textbf{16} & \textbf{red-agent-1} & \textbf{moveDown} & \textbf{{[}0.0,0.0{]}} & \textbf{{[}1.0,1.0{]}} & \textbf{m\_Down\_on} \\ \midrule
17 & red-agent-1 & moveDown & {[}1.0,1.0{]} & {[}0.0,0.0{]} & m\_Down\_off \\
17 & (red-agent-1,16) & atLoc & {[}0.0,1.0{]} & {[}1.0,1.0{]} & m\_Set\_location \\
17 & (red-agent-1,24) & atLoc & {[}1.0,1.0{]} & {[}0.0,0.0{]} & m\_Rem\_location \\
17 & red-agent-1 & moveDown & {[}0.0,0.0{]} & {[}1.0,1.0{]} & m\_Down\_on \\ \midrule
18 & red-agent-1 & moveDown & {[}1.0,1.0{]} & {[}0.0,0.0{]} & m\_Down\_off \\
18 & (red-agent-1,8) & atLoc & {[}0.0,1.0{]} & {[}1.0,1.0{]} & m\_Set\_location \\
18 & (red-agent-1,16) & atLoc & {[}1.0,1.0{]} & {[}0.0,0.0{]} & m\_Rem\_location \\
18 & blue-agent-1 & shootLeftB & {[}0.0,1.0{]} & {[}1.0,1.0{]} & s\_Left\_on \\
18 & (blue-bullet-1,3) & atLoc & {[}0.0,1.0{]} & {[}1.0,1.0{]} & s\_Set\_location \\
18 & (blue-bullet-1,left) & direction & {[}0.0,1.0{]} & {[}1.0,1.0{]} & s\_Set\_dir \\
18 & red-agent-1 & moveDown & {[}0.0,0.0{]} & {[}1.0,1.0{]} & m\_Down\_on \\ \midrule
19 & red-agent-1 & moveDown & {[}1.0,1.0{]} & {[}0.0,0.0{]} & m\_Down\_off \\
19 & (red-agent-1,0) & atLoc & {[}0.0,1.0{]} & {[}1.0,1.0{]} & m\_Set\_location \\
19 & (red-agent-1,8) & atLoc & {[}1.0,1.0{]} & {[}0.0,0.0{]} & m\_Rem\_location \\
19 & blue-agent-1 & shootLeftB & {[}1.0,1.0{]} & {[}0.0,0.0{]} & s\_Left\_off \\
19 & (blue-bullet-1,3) & atLoc & {[}1.0,1.0{]} & {[}0.0,0.0{]} & s\_Rem\_location \\
19 & (blue-bullet-1,2) & atLoc & {[}0.0,1.0{]} & {[}1.0,1.0{]} & s\_Set\_location \\
19 & red-agent-1 & moveUp & {[}0.0,0.0{]} & {[}1.0,1.0{]} & m\_Up\_on \\ \midrule
20 & red-agent-1 & moveUp & {[}1.0,1.0{]} & {[}0.0,0.0{]} & m\_Up\_off \\
20 & (red-agent-1,8) & atLoc & {[}0.0,0.0{]} & {[}1.0,1.0{]} & m\_Set\_location \\
20 & (red-agent-1,0) & atLoc & {[}1.0,1.0{]} & {[}0.0,0.0{]} & m\_Rem\_location \\
20 & (blue-bullet-1,2) & atLoc & {[}1.0,1.0{]} & {[}0.0,0.0{]} & s\_Rem\_location \\
20 & (blue-bullet-1,1) & atLoc & {[}0.0,1.0{]} & {[}1.0,1.0{]} & s\_Set\_location \\
20 & red-agent-1 & moveUp & {[}0.0,0.0{]} & {[}1.0,1.0{]} & m\_Up\_on \\ \midrule
21 & red-agent-1 & moveUp & {[}1.0,1.0{]} & {[}0.0,0.0{]} & m\_Up\_off \\
21 & (red-agent-1,16) & atLoc & {[}0.0,0.0{]} & {[}1.0,1.0{]} & m\_Set\_location \\
21 & (red-agent-1,8) & atLoc & {[}1.0,1.0{]} & {[}0.0,0.0{]} & m\_Rem\_location \\
21 & (blue-bullet-1,1) & atLoc & {[}1.0,1.0{]} & {[}0.0,0.0{]} & s\_Rem\_location \\
21 & (blue-bullet-1,0) & atLoc & {[}0.0,1.0{]} & {[}1.0,1.0{]} & s\_Set\_location \\
\bottomrule
\end{tabular}
\end{table}

\section{Related Work}
\label{sec:related}
A lifelong learner AI suggested in~\cite{nirenburg2023hybrid} starts from a hand-crafted knowledge base in the form of symbolic rules and then employs deep learning techniques to grow its knowledge base through experience. Due to costs, risk, reliability and availability of real life data, such experience is often gained using simulators. PyReason, designed to support logically defined environments, qualifies as an ideal candidate for emerging AI agents of this kind. It is to be noted that temporal logic programming is different from temporal logic. The main difference being that temporal logic relies (typically) on an MDP as the underling structure and the rules are just used for specification checking (shielding can be viewed as an application of this). We use temporal logic programming~\cite{dekhtyar1999temporal}, which is the notion of a collection of temporal logic rules to specify the environmental dynamics. Another thing to note here is that, Portability and Transfer are different. Transfer learning in RL~\cite{zhu2023transfer} involves leveraging knowledge gained from one task or environment to improve learning and performance on a related but different task or environment. What we show here is portability whereby we leverage a fast, scalable simulation environment in PyReason to learn policies which are then used for an identical task in a slower simulation environment which would have been prohibitively slow to carry out the same number of training epochs. Although the slower simulator models the same environment, it may lack several of PyReason's capabilities like explainability and logical shielding. Like our approach, hierarchical reinforcement learning (HRL)~\cite{pateria2021hierarchical} also offers a semantic coarsening to improve agent performance.  However, unlike our approach, HRL coarsens the action space by creating a hierarchy - where our approach is coarsening the environment itself.  As our approach is agnostic to the RL training regime, HRL and our approach are actually complementary and a represent a promising avenue for future research.

\section{Conclusions and Future Work}
\label{sec:future}
In this paper we presented a logic-based semantic proxy for the simulator in an RL pipeline.  We attained significant speedup while providing comparable agent performance. While the policy produced by our approach can be considered as a a set of rules, the rule bodies consist of all ground atoms - hence we seek to leverage frameworks such as~\cite{delfosse2023interpretable} to produce more compact policy rules.  Another area of exploration is the use of this framework to identify issues relating to the sim-to-real gap.  Finally, the description of the environment using natural language is also an area that can be explored due to recent advances in translating natural language to temporal logic formulas using LLMs~\cite{liu2022lang2ltl}.

\section*{Acknowledgments}
\noindent Some of the authors were funded by the Arizona New Economic Initiative MADE STC as well as funding from SSCI.

\bibliographystyle{IEEEtran}
\bibliography{pyreason-rl}

\end{document}